\documentclass[nohyperref]{article}

\usepackage{microtype}
\usepackage{graphicx}
\usepackage{subfigure}
\usepackage{booktabs} 

\usepackage{hyperref}

\usepackage[accepted]{icml2022}

\usepackage{amsmath}
\usepackage{amssymb}
\usepackage{mathtools}
\usepackage{amsthm}

\usepackage{latexsym}

\usepackage{color}

\def\G{\mathcal{G}}

\def\D{\mathcal{D}}

\def\our{HyperNeRFGAN}

\usepackage[capitalize,noabbrev]{cleveref}

\theoremstyle{plain}

\theoremstyle{definition}

\theoremstyle{remark}

\usepackage[textsize=tiny]{todonotes}

\begin{document}

\twocolumn[
\icmltitle{\our{}: Hypernetwork approach to 3D NeRF GAN}




\begin{icmlauthorlist}
\icmlauthor{Adam Kania}{yyy}
\icmlauthor{Artur Kasymov}{yyy}
\icmlauthor{Jakub Kościukiewicz}{yyy}
\icmlauthor{Artur Górak}{yyy}
\icmlauthor{Marcin Mazur}{yyy}
\icmlauthor{Maciej Zieba}{comp}
\icmlauthor{Przemys{\l}aw Spurek}{yyy}
\end{icmlauthorlist}

\icmlaffiliation{yyy}{Faculty of Mathematics and Computer Science, Jagiellonian University 6 Lojasiewicza Street, 30-348 Kraków, Poland}
\icmlaffiliation{comp}{Department of Artificial Intelligence, University of Science and Technology Wyb. Wyspianskiego 27, 50-370, Wrocław, Poland}

\icmlcorrespondingauthor{Przemys{\l}aw Spurek}{przemyslaw.spurek@uj.edu.pl}

\icmlkeywords{NeRF, GAN}

\vskip 0.3in
]



\printAffiliationsAndNotice{\icmlEqualContribution} 

\begin{abstract}

The recent surge in popularity of deep generative models for 3D objects has highlighted the need for more efficient training methods, particularly given the difficulties associated with training with conventional 3D representations, such as voxels or point clouds. Neural Radiance Fields (NeRFs), which provide the current benchmark in terms of quality for the generation of novel views of complex 3D scenes from a limited set of 2D images, represent a promising solution to this challenge. However, the training of these models requires the knowledge of the respective camera positions from which the images were viewed. In this paper, we overcome this limitation by introducing \our{}, a Generative Adversarial Network (GAN) architecture employing a hypernetwork paradigm to transform a Gaussian noise into the weights of a NeRF architecture that does not utilize viewing directions in its training phase. Consequently, as evidenced by the findings of our experimental study, the proposed model, despite its notable simplicity in comparison to existing state-of-the-art alternatives, demonstrates superior performance on a diverse range of image datasets where camera position estimation is challenging, particularly in the context of medical data.

\end{abstract}

\section{Introduction}
\label{intro}


Generative Adversarial Networks (GANs)~\cite{goodfellow2014generative} facilitate the generation of high-quality 2D images~\cite{yu2017unsupervised,karras2017progressive,karras2019style,karras2020analyzing,struski2022locogan}. However, achieving a comparable level of quality for 3D objects remains a challenge. The primary difficulty arises from the necessity of extensive deep architectures in conjunction with the use of 3D representations, including voxels and point clouds, which present challenges in an accurate color rendering. One potential solution to this issue is the direct working within the 2D image domain, as exemplified by Neural Radiance Fields (NeRFs)~\cite{mildenhall2021nerf}, which permit the generation of unseen views of intricate 3D scenes based on a limited set of 2D images. In essence, by leveraging the relationships between these base images and computer graphics techniques such as ray tracing, these neural models are capable of producing high-quality renderings of 3D objects from novel viewpoints.

\begin{figure} 
\begin{center} 
\includegraphics[width=0.9\columnwidth]{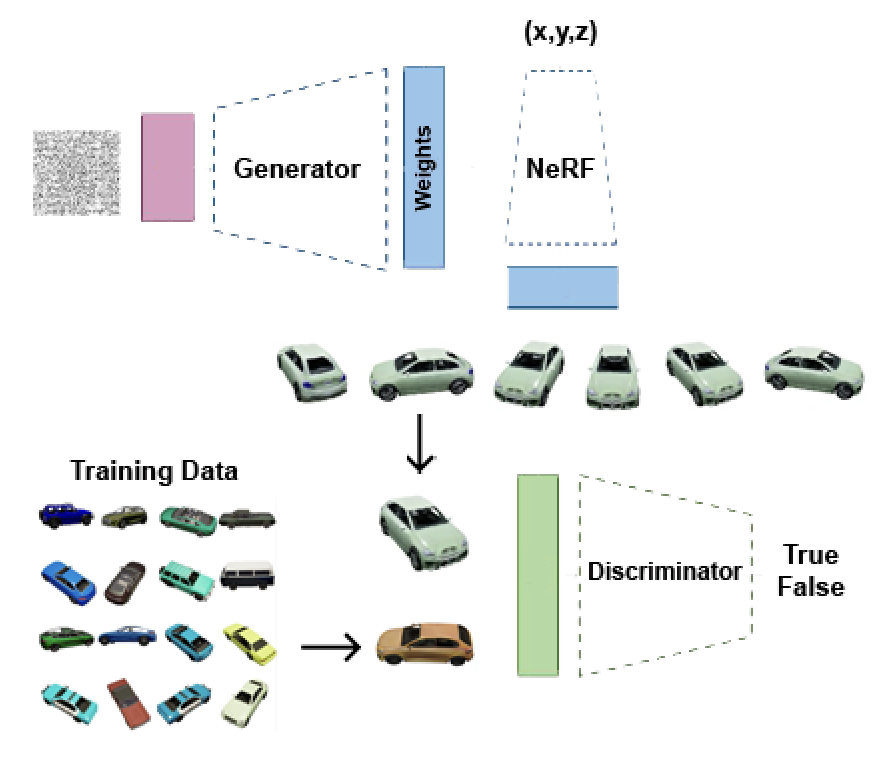} 
\end{center} 
  \caption{Our \our{} model employs a hypernetwork to convert a Gaussian noise into the weights of the simplified NeRF architecture (not requiring information about camera positions), subsequently used for the generation of novel 2D views. During the training phase, a standard GAN-based framework (incorporating a typical 2D discriminator) is employed. Despite the generation of 2D images, our model utilizes a 3D-aware NeRF representation, thereby facilitating precise 3D object generation.} 
\label{fig:teaser} 
\end{figure}

\begin{figure*}
    \centering
    \includegraphics[width=0.9\textwidth]{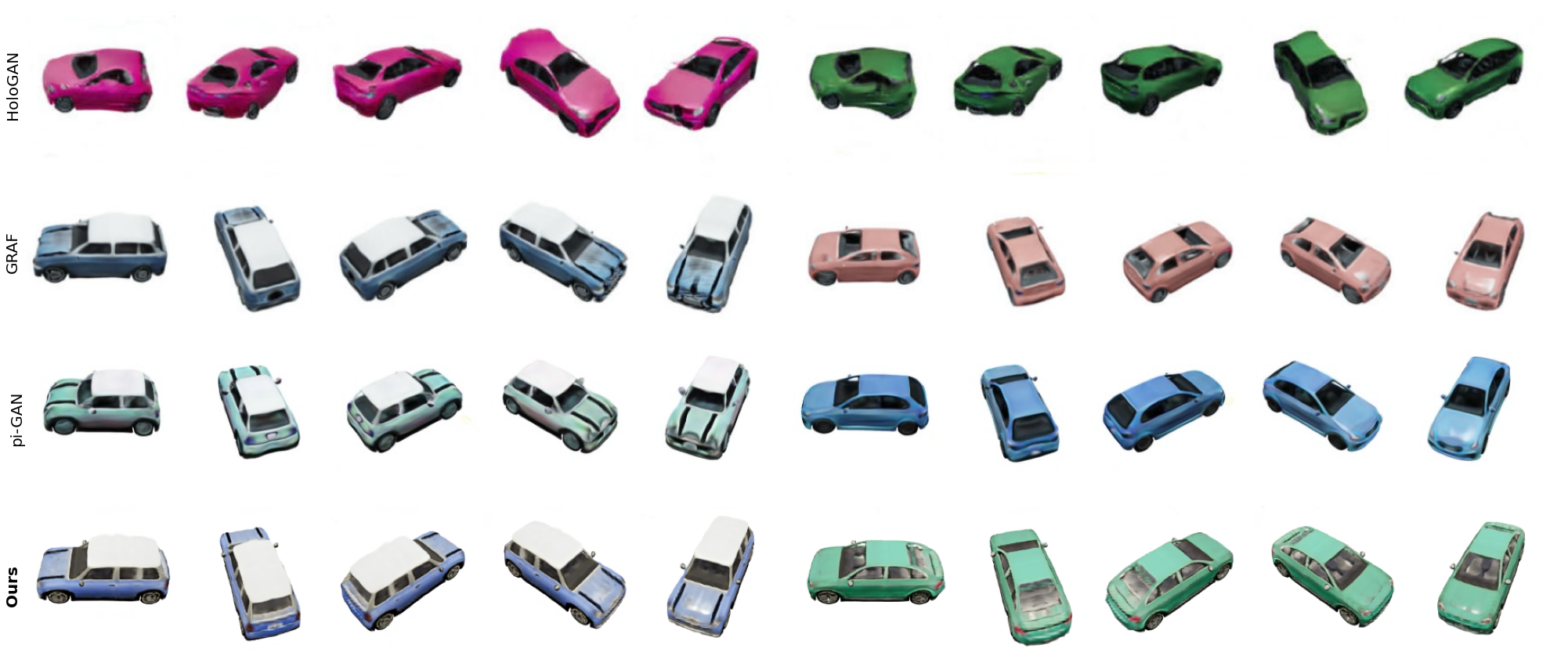}
    \caption{Qualitative comparison of \our{} (our) with HoloGAN~\cite{nguyen2019hologan},  
GRAF~\cite{schwarz2020graf}, and $\pi$-GAN~\cite{chan2021pi} trained on the CARLA dataset~\cite{dosovitskiy2017carla}.  It is noteworthy that our model has been shown to produce outcomes that are comparable to those of the most successful competitor, namely $\pi$-GAN.}
    \label{fig:carla_comparison}
\end{figure*}

It must be acknowledged that the incorporation of NeRF representation with a GAN architecture is not a straightforward process due to the inherent complexity of the NeRF conditioning mechanism~\cite{rebain2022attention}. Consequently, the majority of models tend to favour the use of the SIREN~\cite{sitzmann2020implicit} architecture instead, as this allows for natural conditioning. However, the quality of rendered 3D objects is somewhat inferior when NeRF is replaced with SIREN, as exemplified by models such as GRAF~\cite{schwarz2020graf} and $\pi$-GAN~\cite{chan2021pi} which employ SIREN along with a conditioning mechanism to generate implicit representations. While these approaches yield promising results, the substitution of SIREN with NeRF in such frameworks presents an intriguing yet complex issue.


In order to address the aforementioned challenge, this paper incorporates NeRF and GAN architectures by means of the hypernetwork paradigm~\cite{ha2016hypernetworks}. Specifically, we introduce the \our{} model, which employs a hypernetwork designed to convert a Gaussian noise into the weights of a simplified NeRF target architecture that no longer utilizes viewing directions to produce the output for given 3D positions. Subsequently, the novel 2D views rendered by NeRF are evaluated by a traditional 2D discriminator. Consequently, the entire model is implicitly trained in accordance with the principles of a GAN-based framework. A diagrammatic representation of the proposed architectural design is presented in Figure \ref{fig:teaser}. It should be noted that although \our{} generates 2D images, it employs a 3D-aware NeRF representation, thereby ensuring that the model produces accurate 3D objects. Furthermore, as the model does not employ viewing directions during training, it can be successfully applied to a variety of datasets where camera position estimation may be challenging or impossible.
To illustrate this, we apply \our{} to three different datasets: CARLA~\cite{dosovitskiy2017carla} and CelebA~\cite{liu2015deep}, which are both real-world datasets, as well as a medical dataset consisting of digitally reconstructed radiographs (DRR) of chests and knees~\cite{coronafigueroa2022mednerf}. The results of the experiments conducted demonstrate that our solution, despite its notable simplicity, outperforms (or is at least on a similar level to) existing state-of-the-art methods, including HoloGAN~\cite{nguyen2019hologan},  
GRAF~\cite{schwarz2020graf}, and $\pi$-GAN~\cite{chan2021pi} for CARLA and CelebA, and GRAF~\cite{schwarz2020graf}, pixelNeRF~\cite{yu2021pixelnerf}, MedNeRF~\cite{coronafigueroa2022mednerf}, and UMedNeRF~\cite{hu2024umednerf} for the medical dataset.


In conclusion, our contribution can be summarized as follows:
\begin{itemize}
    \item we introduce \our{}, an implicit GAN-based model that generates a simplified 3D-aware NeRF representation (not requiring information about camera positions) directly through the use of a hypernetwork,
    \item the derived NeRF architecture is employed for the generation of novel 2D views, subsequently evaluated through a conventional 2D discriminator (thus, the complete model is trained in accordance with the tenets of a 2D adversarial methodology),
    \item the experiments conducted demonstrate that \our{}, despite its notable simplicity in comparison to existing state-of-the-art alternatives, exhibits superior performance on a diverse range of image datasets where camera position estimation is challenging, particularly in the context of medical data.
\end{itemize}

\section{Related work}

Generative Adversarial Networks (GANs)~\cite{goodfellow2014generative} permit the generation of high-quality images~\cite{yu2017unsupervised,karras2017progressive,karras2019style,karras2020analyzing,struski2022locogan}. However, GANs operate on 2D dimensional images and thus fail to account for the 3D dimensional nature of the physical world.

\begin{figure}
    \centering
    \includegraphics[width=0.48\textwidth]{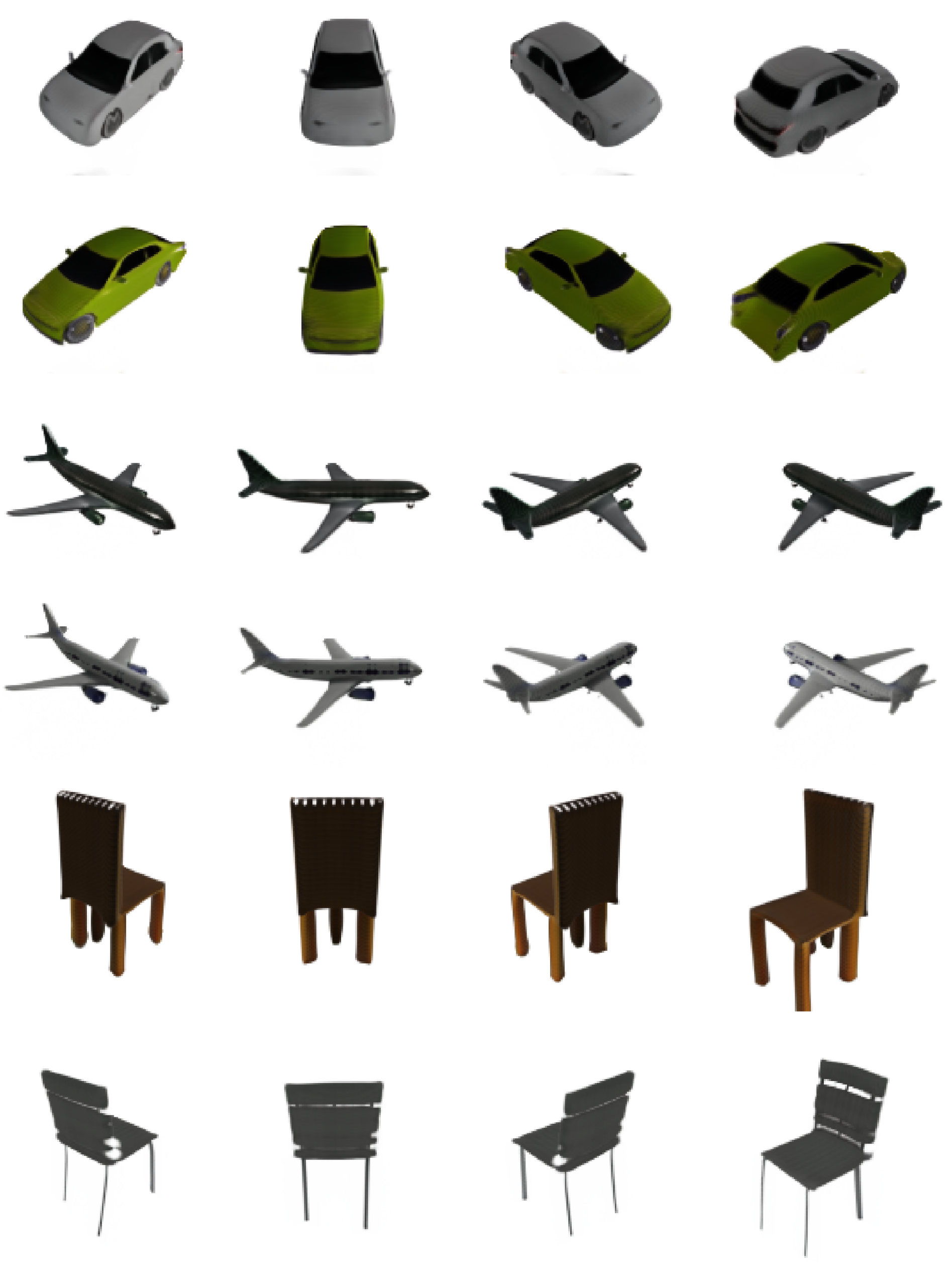}
    \caption{Sample 2D images generated by the \our{} model (our) trained on the ShapeNet-based dataset proposed in~\cite{zimny2022points2nerf}, consisting of 50 images of each object from the car, plane, and chair classes.}
    \label{fig:shapenet_examples}
\end{figure}

The initial approaches for 3D-aware image synthesis were Visual Object Network~\cite{zhu2018visual} and PrGAN~\cite{gadelha20173d}. These methods generate voxelized 3D shapes using a 3D-GAN~\cite{wu2016learning} and then project them into 2D.
An alternative approach is taken by HoloGAN~\cite{nguyen2019hologan} and BlockGAN~\cite{nguyen2020blockgan}, which employ a similar fusion but utilize implicit 3D representation to model a 3D representation of the world. The use of an explicit volume representation has, however, constrained the resolution of the models, as discussed in~\cite{lunz2020inverse}. Furthermore, the authors of~\cite{szabo2019unsupervised} suggest the use of meshes to represent 3D geometry. Conversely, in~\cite{liao2020towards}, collections of primitives for image synthesis are employed.

\begin{figure*}[htb]
    \centering
    \includegraphics[width=0.9\textwidth]{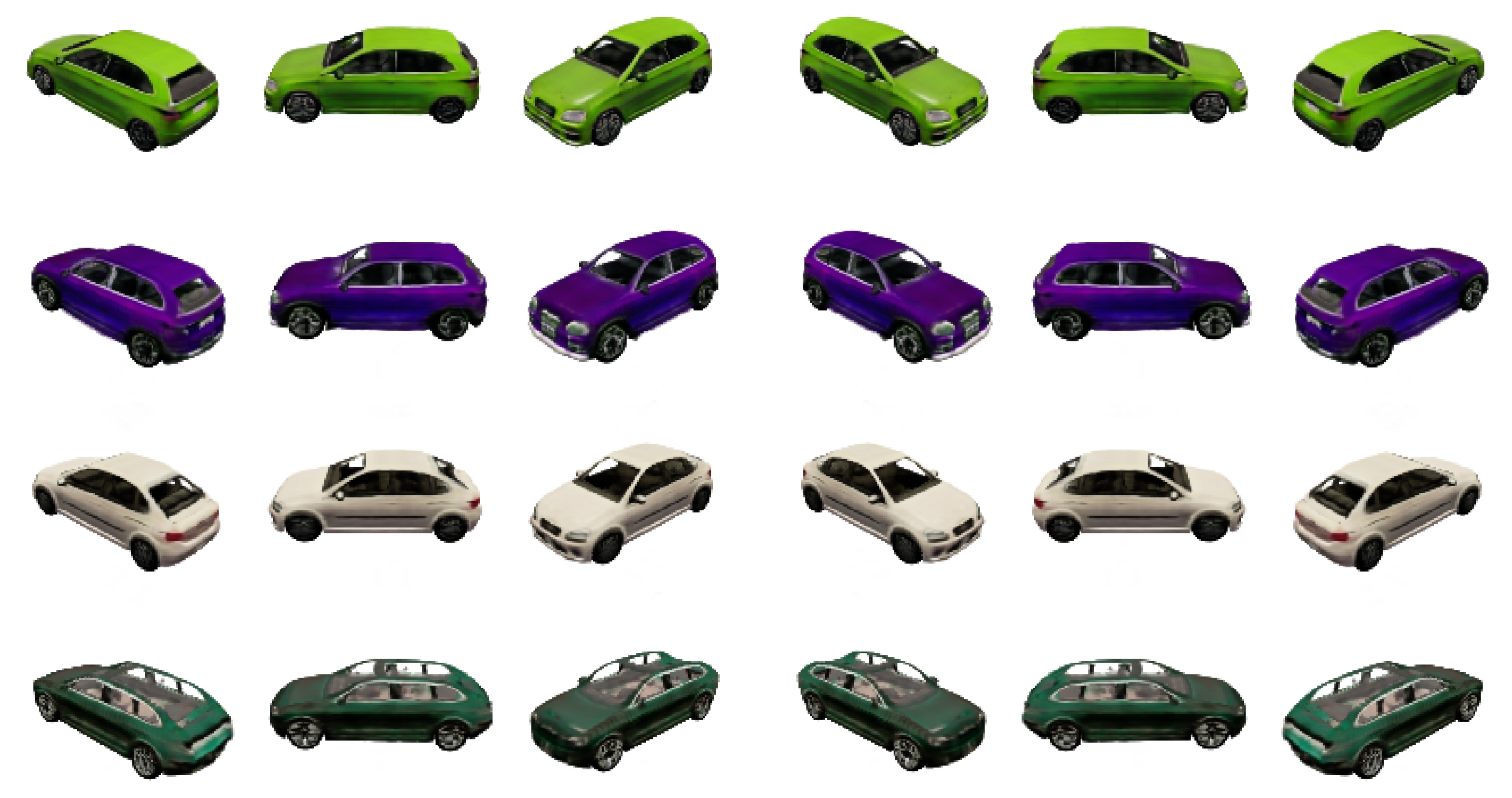}
    \caption{Sample 2D images generated by the \our{} model (our) trained on the CARLA dataset~\cite{dosovitskiy2017carla}. It should be noted that our method permits the effective modeling of transparency in car windows.}
    \label{fig:carla_examples}
\end{figure*}

GRAF~\cite{schwarz2020graf} and $\pi$-GAN~\cite{chan2021pi} employ implicit neural radiance fields for the generation of 3D-aware images and geometry. These models also utilize the SIREN algorithm in conjunction with a conditioning mechanism.
As an alternative option, ShadeGAN~\cite{pan2021shading} employs a shading-guided pipeline, whereas GOF~\cite{xu2021generative} utilizes a gradual reduction in the sampling region of each camera ray. In the GIRAFFE approach~\cite{niemeyer2021giraffe}, the initial step is to generate low-resolution feature maps. In the second step, the representation is passed to a 2D convolutional neural network (CNN) to generate outputs at a higher resolution.
In contrast, StyleSDF~\cite{or2022stylesdf} integrates an SDF-based 3D representation with a StyleGAN2 for image generation, whereas in~\cite{chan2022efficient}, the authors employ a StyleGAN2 generator and a tri-plane representation of 3D objects. These approaches yield superior results in terms of generated object quality but are exceedingly challenging to train.

A distinct family of models has been developed for use in medical applications. One such model is MedNeRF~\cite{coronafigueroa2022mednerf}, which employs a GRAF-based approach to generate precise 3D projections from a single X-ray view. By combining GRAF with DAG~\cite{Tran_2021} and employing multiple discriminator heads, the authors of~\cite{coronafigueroa2022mednerf} achieved markedly superior results compared to standard GRAF on the medical dataset. Another related model, UMedNeRF~\cite{hu2024umednerf}, is constructed upon MedNeRF. It employs automated calculation of weight parameters for discriminator losses, thereby facilitating more precise adaptation to the specific requirements of each task. This results in enhanced image clarity and the discernment of finer details within bone structures, when compared to its underlying MedNeRF architecture.

\section{\our{}: hypernetwork for generating NeRF representions}
\label{sec:method}

This section presents the \our{} model. The fundamental premise of the proposed GAN-based methodology is that the generator serves as a hypernetwork, transforming a noise vector, sampled from a Gaussian distribution, into the weights of a NeRF representation of a given 3D object. Consequently, it is possible to generate a multitude of images of the object from a variety of perspectives in a manner that is fully controllable. Furthermore, the use of NeRF-based image rendering allows the discriminator to operate on generated 2D images, which is a notable simplification compared to existing state-of-the-art GAN-based models, such as HoloGAN~\cite{nguyen2019hologan}, GRAF~\cite{schwarz2020graf}, and $\pi$-GAN~\cite{chan2021pi}, which are fed by complex 3D structures. We begin by outlining the fundamental concepts that underpin our approach, after which we proceed to present the architectural and training details of the \our{} model.

\paragraph{Hypernetworks}

As defined in \cite{ha2016hypernetworks}, hypernetworks are neural models that generate weights for another target network with the objective of solving a specific task. This approach results in a reduction of the number of trainable parameters in comparison to traditional methodologies that integrate supplementary information into the target model via a single embedding. A notable reduction in the size of the target model is achievable due to the fact that it does not share global weights. Instead, these weights are provided by the hypernetwork. Similarly, the authors of \cite{sheikh2017stochastic} employ this mechanism to generate a variety of target networks that approximate the same function, thereby establishing a parallel between hypernetworks and generative models.

Hypernetworks have a multitude of applications, including few-shot learning \cite{sendera2023hypershot} and probabilistic regression \cite{zikeba2020regflow}. Additionally, they are utilized in numerous techniques to generate continuous 3D object representations \cite{spurek2020hypernetwork,spurek2021general}. For instance, HyperCloud~\cite{spurek2020hypernetwork} employs a classical MLP as the target model to transform points from a uniform distribution on the unit sphere into point clouds that conform to the desired shape. Conversely, in \cite{spurek2021general}, the target model is a Continuous Normalizing Flow~\cite{grathwohl2018ffjord}, a generative model that constructs the point cloud from an assumed base distribution in 3D space.

\paragraph{NeRF representation of 3D objects}

A Neural Radiance Field (NeRF) is a scene modeling technique that uses a fully-connected network to represent the visual data \cite{mildenhall2021nerf}. The input to NeRF is a 5D coordinate, comprising a spatial position ${\bf x}=(x, y, z)$ and a viewing direction ${\bf d}=(\theta,\psi)$. The output is an emitted color ${\bf c}=(r, g, b)$ and a volume density $\sigma$.

A standard NeRF model is trained using a collection of images. In this context, numerous rays are generated that intersect both the image and a 3D object, which is modeled by a multi-layer perceptron (MLP) network 
$$ 
F_{\theta} \colon ({\bf x}, {\bf d}) \to ( {\bf c} , \sigma) 
$$ 
with parameters $\theta$ adjusted to map each 5D input coordinate to its respective directional color emission and volume density.

The NeRF loss function draws inspiration from traditional volume rendering as described in \cite{kajiya1984ray}. The color is calculated for each ray traversing the scene. The volume density $\sigma=\sigma({\bf x})$ can be regarded as the differential probability of a ray. The anticipated color ${\bf c}={\bf c}({\bf x},\sigma)$ is interpreted as the color of a corresponding camera ray ${\bf r}(t) = {\bf o} + t {\bf d}$ (where ${\bf o}$ is the ray's origin and ${\bf d}$ is its direction). It is determined by an integral and numerically approximated by a sum in practical applications.

\paragraph{Hypernetwork paradigm for generative modeling}

The concept of integrating hypernetworks and deep generative models is a relatively well-established area of research. In~\cite{ratzlaff2019hypergan,henning2018approximating}, the authors construct a GAN to generate the parameters of a neural network dedicated to regression or classification tasks. Other related examples include HyperVAE~\cite{nguyen2020hypervae}, which is designed to encode an arbitrary target distribution by generating parameters of a deep generative model given distribution samples, Point2NeRF~\cite{zimny2022points2nerf}, which uses hypernetwork to output weights of an autoencoder based generative model (VAE) to create NeRF representation from a 3D point cloud, and HCNAF~\cite{oechsle2019texture}, which is a hypernetwork that produces parameters for a conditional auto-regressive flow model \cite{kingma2016improved,oord2018parallel,huang2018neural}. Alternatively, the authors of \cite{skorokhodov2021adversarial} propose INR-GAN, which employs a hypernetwork to generate a continuous representation of images. The hypernetwork is capable of modifying the shared weights through a low-cost mechanism known as factorized multiplicative modulation.

\paragraph{\our{}}

The proposed \our{} utilizes a hypenetwork to produce weights for the NeRF target network. The model adheres to the design patterns of INR-GAN, employing the StyleGAN2 backbone architecture. Consequently, it is trained using the StyleGAN2 objective in a manner analogous to that employed in INR-GAN. Specifically, in each training iteration, the noise vector from the assumed base (Gaussian) distribution  $P_{noise}$ is sampled and transformed using the generator $\G$ to obtain the weights of the target NeRF model $F_{\theta}$. Furthermore, the target model is employed to render 2D images from a variety of perspectives. The generator $\G$ is responsible for creating a 3D representation that enables the generation of 2D images that will be indistinguishable from those created by the true data distribution $P_{data}$. In contrast, the discriminator $\D$ is designed to distinguish between fake renderings and authentic 2D images drawn from the data distribution. Formally, this minimax game is given by the following expression:
\begin{equation}\label{eq:minmax}
	\min_{\G} \max_{\D}  V(\D,\G),
\end{equation}
where 
\begin{equation}\label{eq:objective}
	V =
	\mathbb{E}_{x \sim P_{data}} \log \D(x) +  \mathbb{E}_{z \sim P_{noise}} \log (1-\D(\G(z))).
\end{equation}
A diagrammatic summary of the architectural structure of \our{} is presented in Figure~\ref{fig:teaser}. 

\begin{figure*}
    \centering
    \includegraphics[width=1\textwidth]{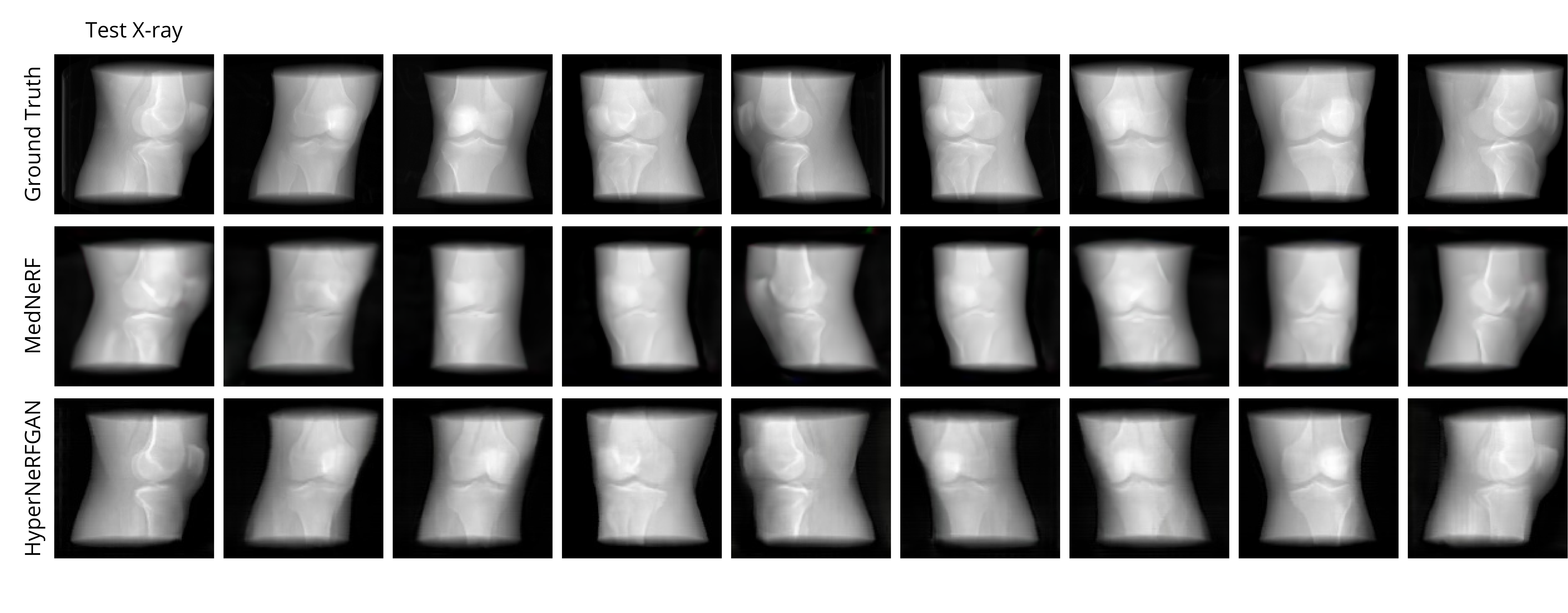}
    \includegraphics[width=.8\textwidth]{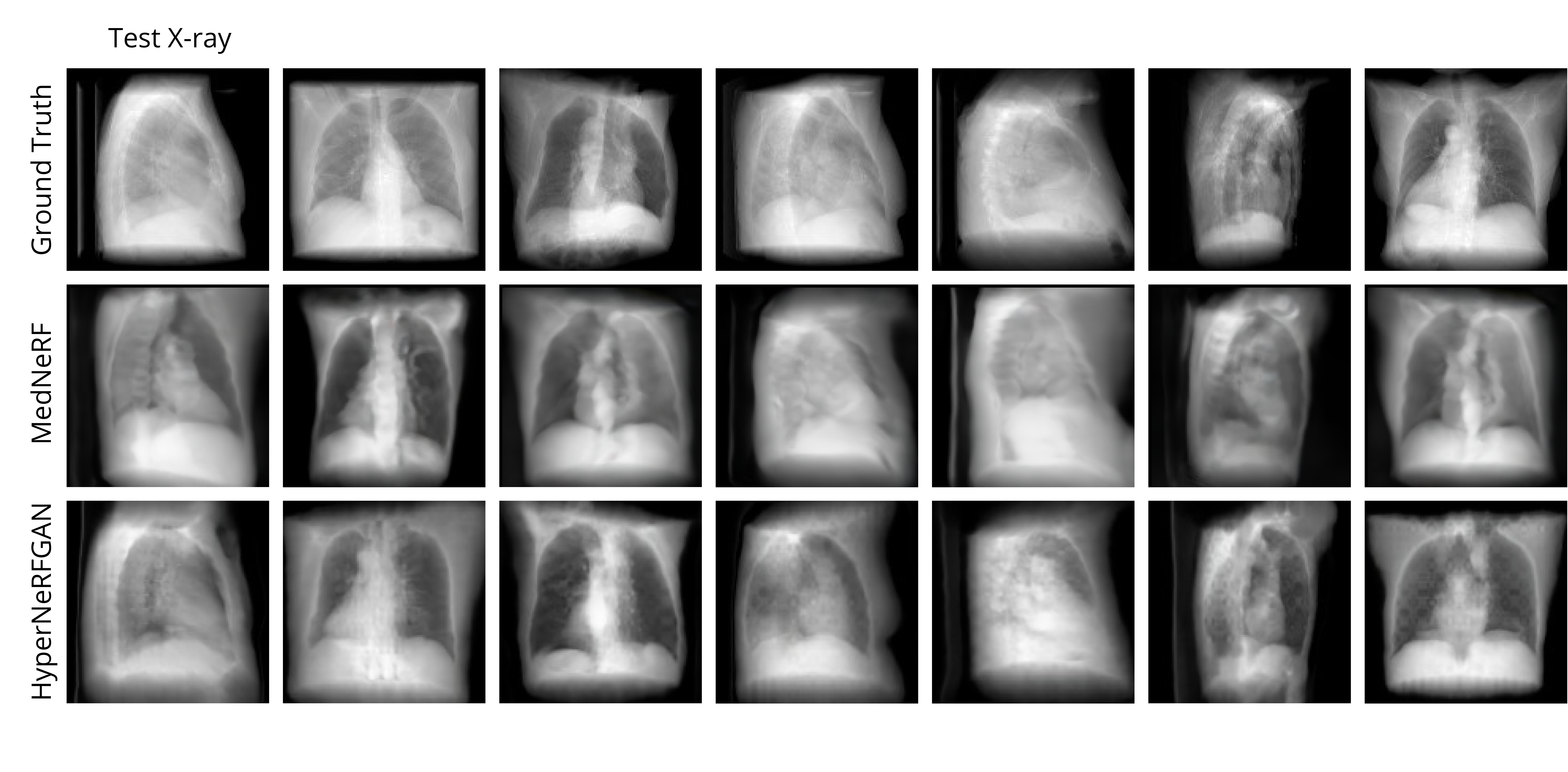}    
        \caption{Qualitative comparison between \our{} (our) and MedNeRF trained on the medical dataset consisting of digitally reconstructed radiographs (DRR) of knees and chests~\cite{coronafigueroa2022mednerf}. Note that our qualitative comparison shows that our model shows a significant improvement in the quality of reconstruction of CT projections.}
        \label{fig:learning}
\end{figure*}

The key distinction between our proposed model and existing state-of-the-art solutions in the field is the use of a special 3D-aware NeRF representation $F_{\theta}$ (instead of SIREN) that differs from the original NeRF in a few aspects. Firstly, in contrast to the standard linear architecture, $F_{\theta}$ employs factorized multiplicative modulation (FMM) layers, as seen in INR-GAN. The FMM layer with an input of size $n_{in}$ and an output of size $n_{out}$ can be defined as follows:
\begin{equation}\label{eq:fmm}
    y = W \odot (A \times B) \cdot x_{in} + b = \Tilde{W} \cdot x_{in} + b,   
\end{equation}
where $W$ and $b$ are matrices that share the parameters across 3D representations, while $A$ and $B$ are two modulation matrices (created by the generator $\G$) with dimensions $n_{out} \times k$ and $k \times n_{in}$, respectively. The parameter $k$ controls the rank of $A \times B$ and exerts an influence on the expressiveness and memory usage. In this regard, higher values of $k$ result in an increase in the expressiveness of the FMM layer, but also in an increase in the amount of memory required by the hypernetwork. In our approach, we always use $k=10$.

Secondly, to reduce the computational expense of training, we do not optimize two networks as in the original NeRF. Instead, we reject the larger ``fine'' network and employ only the smaller ``coarse'' network. Additionally, we reduce the size of the ``coarse'' network by decreasing the number of channels in each hidden layer from 256 to 128. In some experiments, we also decrease the number of layers from 8 to 4.

\begin{figure}
    \centering
    \includegraphics[width=0.49\textwidth]{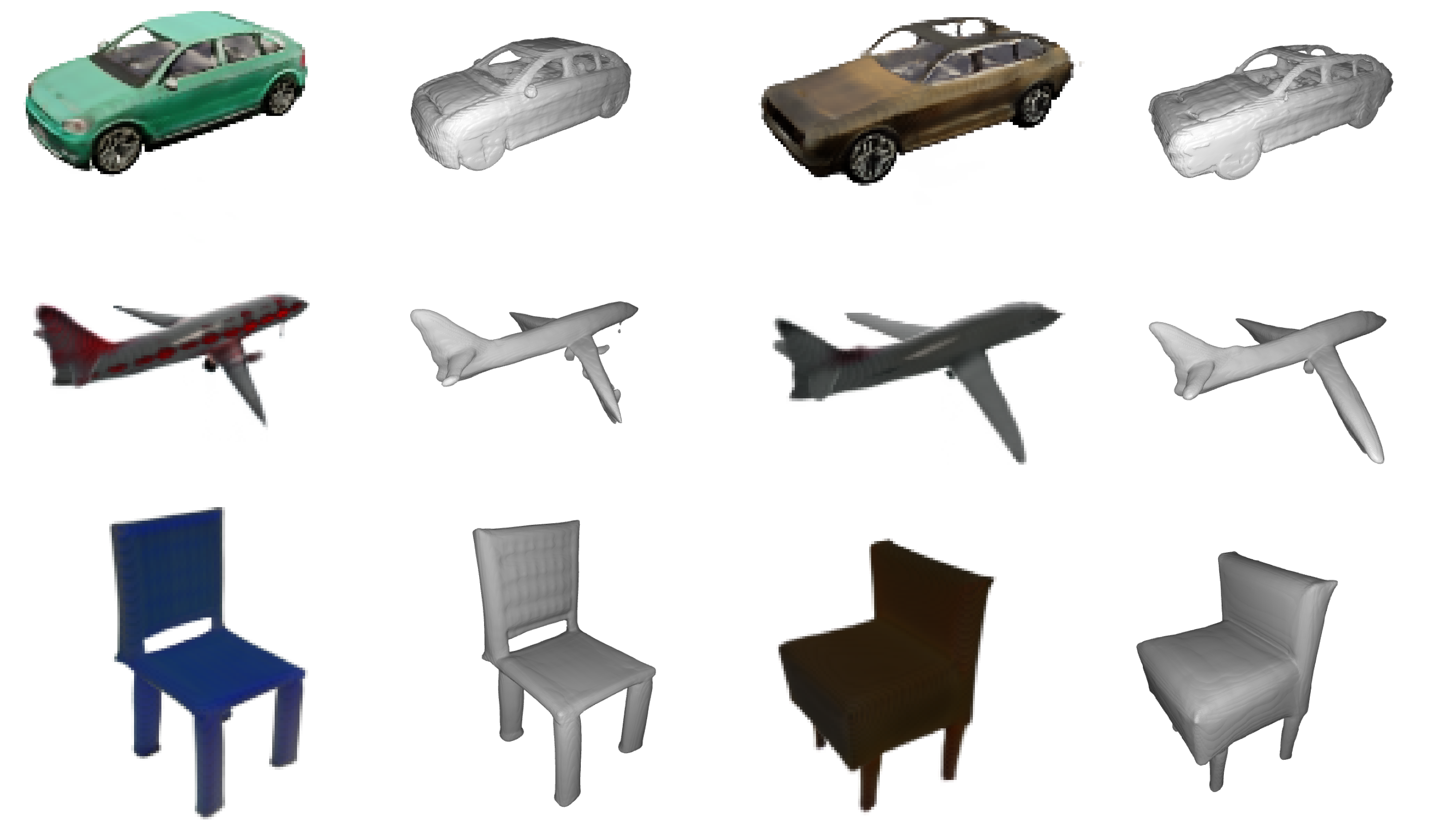}
    \caption{Sample meshes generated by the \our{} model (our) trained on the CARLA dataset~\cite{dosovitskiy2017carla} and two classes (car and plane) from the ShapeNet dataset~\cite{zimny2022points2nerf}.}
    \label{fig:mesh_examples}
\end{figure}

Finally, in contrast to the standard NeRF architecture, our approach does not utilize the viewing direction. Instead, our NeRF representation is a single MLP that takes the spatial location ${\bf x} = (x, y, z)$ and transforms it to the emitted color ${\bf c} = {\bf c}({\bf x})$ and volume density $\sigma=\sigma({\bf x})$, i.e.:
\begin{equation}
    F_{\theta} \colon {\bf x} \to ( {\bf c} , \sigma).
\end{equation}
This is due to the fact that the images utilized for training lack view-dependent characteristics, such as reflections. (However, while our solution does not currently employ viewing direction data, there is no inherent reason to prohibit its use in datasets that would benefit from this additional information.) It should be noted that the use of such an architectural approach allows for the training of our model on a variety of data types that do not provide access to the camera position associated with images or to a substantial number of images for a single object. (In fact, the use of a single unlabeled view per object is sufficient for training.) This allows for the training of our model on medical data, as demonstrated in the following section.

\begin{figure*}[htb]
    \centering
    \includegraphics[width=0.9\textwidth]{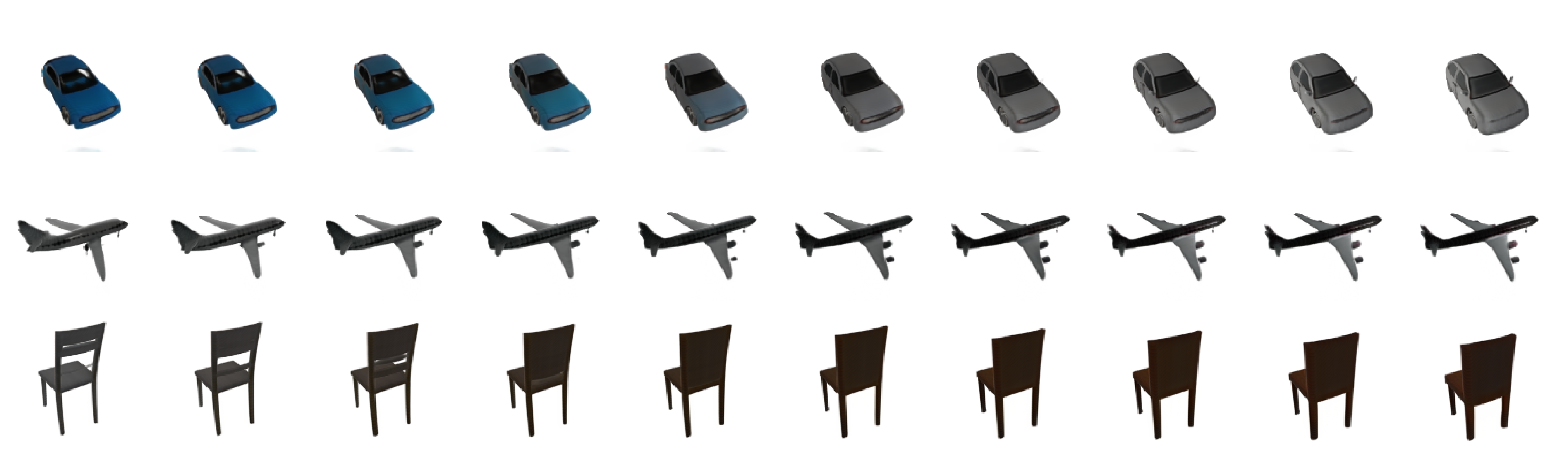}
    \includegraphics[width=0.9\textwidth]{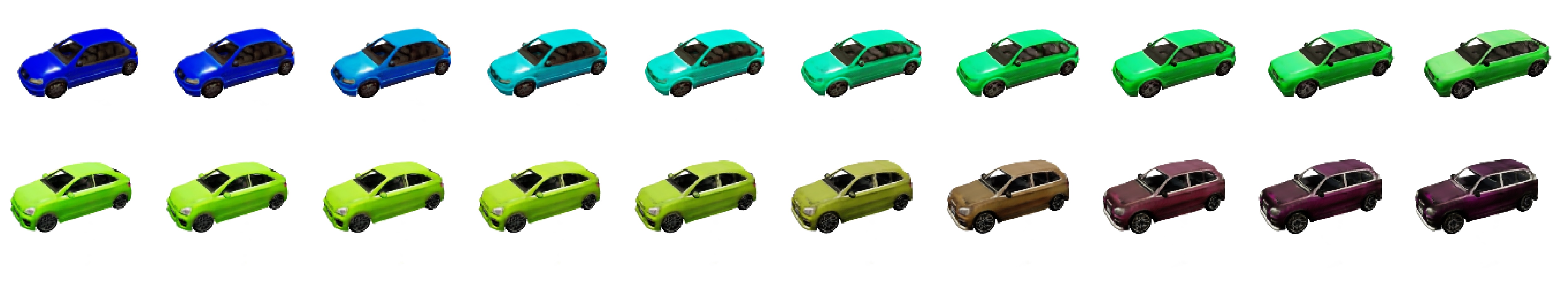}
    \caption{Linear interpolation examples generated by the \our{} model (our) trained on the ShapeNet-based dataset proposed in~\cite{zimny2022points2nerf} (first three rows) and the CARLA dataset~\cite{dosovitskiy2017carla} (last two rows).}
    \label{fig:extra_interpolation}
\end{figure*}

\begin{table*}
    \centering
    \begin{tabular}{ccccc}
    \toprule
    & \multicolumn{2}{c}{Chest dataset } & \multicolumn{2}{c}{Knee dataset} \\
    & \(\downarrow\) FID (\(\mu\) \(\pm\) \(\sigma\))  & \(\downarrow\) KID (\(\mu\) \(\pm\) \(\sigma\)) & \(\downarrow\) FID (\(\mu\) \(\pm\) \(\sigma\)) & \(\downarrow\) KID (\(\mu\) \(\pm\) \(\sigma\)) \\
    \midrule
             GRAF & 68.25 $\pm$ 0.954 & 0.053 $\pm$ 0.0008 & 76.70 $\pm$ 0.302 & 0.058 $\pm$ 0.0001 \\
             pixelNeRF & 112.96 $\pm$ 2.356 & 0.084 $\pm$ 0.0012 & 166.40 $\pm$ 2.153 & 0.158 $\pm$ 0.0010 \\
             MedNeRF & 60.26 $\pm$ 0.322 & \bf 0.041 $\pm$ 0.0005 & 76.12 $\pm$ 0.193 & 0.052 $\pm$ 0.0004\\
             UMedNeRF & 60.25 $\pm$ 0.642 & 0.043 $\pm$ 0.0011 & 70.73 $\pm$ 1.665 & \bf 0.041 $\pm$ 0.0012 \\
             HyperNeRFGAN & \bf 53.53 $\pm$ 0.917 & 0.043 $\pm$ 0.0015 & \bf 57.65 $\pm$ 0.689 & 0.044 $\pm$ 0.0007\\
        
    \bottomrule
    \end{tabular}
    \caption{Quantitative evaluation of \our{} (our) in comparison with
GRAF~\cite{schwarz2020graf}, and pixelNeRF\cite{yu2021pixelnerf}, MedNeRF~\cite{coronafigueroa2022mednerf}, and UMedNeRF~\cite{hu2024umednerf}, in terms of the FID (lower is better) and KID (lower is better) metrics. trained on the medical dataset consisting of digitally reconstructed radiographs (DRR) of chests and knees~\cite{coronafigueroa2022mednerf}.
All scores are the average of 3 runs.
The obtained results prove the superior performance of our solution with respect to the baseline methods.}
    \label{tab:fid_kid_scores}
    \vspace{1cm}
\end{table*}

\section{Experiments}
\label{exp}
 
In this section, we conduct an empirical evaluation of our \our{} method in comparison to the state-of-the-art solutions. Firstly, we undertake a comparative analysis of the quality of generated 3D objects produced by a range of models trained on two distinct datasets. The first is a dataset comprising 2D images of 3D objects obtained from ShapeNet~\cite{zimny2022points2nerf}, and the second is the CARLA dataset~\cite{dosovitskiy2017carla}, which includes images of cars. It should be noted that these datasets are particularly well-suited to our purposes, as each object is presented from only a few perspectives.  Secondly, the classical CelebA dataset~\cite{liu2015deep}, which contains photographs of faces, is employed to assess the performance of different state-of-the-art generative models designed for 3D-aware image synthesis~\cite{chan2022efficient}. It should be noted that from the perspective of 3D generation, this task presents a significant challenge, given that the only available source data are the photographed front sides of faces. Finally, the effectiveness of our model is evaluated using a dataset comprising digitally reconstructed radiographs (DRR) of chests and knees~\cite{coronafigueroa2022mednerf}, in order to ensure its comparability with existing methods.




\begin{figure}
    \centering
    \our{} \qquad\qquad\qquad\quad $\pi$-GAN \quad\\
    \includegraphics[width=0.49\textwidth]{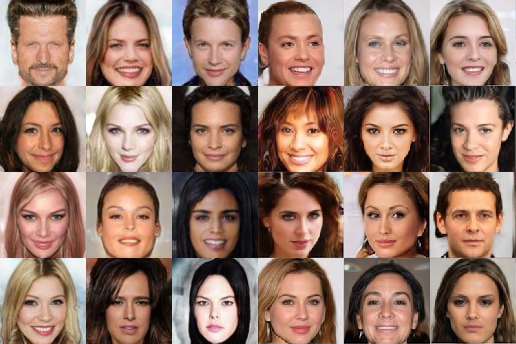}
    \caption{Qualitative comparison between \our{} (our) and $\pi$-GAN trained on the CelebA dataset~\cite{liu2015deep}. It can be seen that both methods demonstrate comparable performance.}
    \label{fig:celeba_examples}
    
\end{figure}

\paragraph{3D object generation}

In the initial experiment, a ShapeNet-based dataset comprising 50 images of each object from the plane, chair, and car categories was utilized. The data were obtained from \cite{zimny2022points2nerf}, where the authors propose the Point2NeRF model, which thus serves as a natural baseline for our method. Figure~\ref{fig:shapenet_examples} presents the 3D objects produced by \our{}, while Figure~\ref{fig:extra_interpolation} additionally displays the results of linear interpolation. It is evident that our approach produces high-quality renders. This is corroborated by the findings of our quantitative study, detailed in Table~\ref{tab:cars_fid}. 

\begin{table}
\centering
\begin{tabular}{cccc}
\toprule
 ShapeNet & Car & Plane & Chair \\
\midrule
Points2NeRF & 82.1 & 239 & 129.3 \\ 
\midrule
\our{} (our) & \bf 29.6 & \bf 33.4 & \bf 22.0 \\
\bottomrule\\
\end{tabular}

\caption{Quantitative evaluation of \our{} (our) in comparison with the autoencoder-based Point2NeRF model~\cite{zimny2022points2nerf} in terms of the FID metric (lower is better). The models were trained on three datasets consisting of 50 images from the car, plane, and chair classes of ShapeNet.
The obtained results clearly demonstrate the superiority of our proposed solution.}
\label{tab:cars_fid}
\end{table}


In the second experiment, we evaluate the performance of our model on the CARLA dataset~\cite{dosovitskiy2017carla} in comparison to other GAN-based models, namely HoloGAN~\cite{nguyen2019hologan},  
GRAF~\cite{schwarz2020graf} and $\pi$-GAN~\cite{chan2021pi}. It should be noted that CARLA comprises only a single image per object (a car), but that we have nonetheless access to photographs of the objects captured from a range of perspectives. The qualitative comparison is presented in Figure~\ref{fig:carla_comparison}, while Table~\ref{tab:carla_fid_kis_is} delivers the results of the quantitative comparison in terms of the Fr\'{e}chet Inception Distance (FID)~\cite{heusel2017gans}, Kernel Inception Distance (KID)~\cite{binkowski2018demystifying}, and Inception Score (IS)~\cite{salimans2016improved}. It is evident that \our{} outperforms all the competitors. Furthermore, as illustrated in Figure~\ref{fig:carla_examples}, our method allows for the effective modeling of transparency in car windows.

It should be noted that the \our{} model, due to its ability to represent NeRF implicitly, can produce high-quality meshes of 3D objects. This is demonstrated in Figure~\ref{fig:mesh_examples}, which presents meshes generated by the CARLA-trained model and two classes of~ShapeNet.

\begin{table}
    \centering
        \begin{tabular}{cccc}
        \toprule
         CARLA & \(\downarrow\) FID & \(\downarrow\) KID$\times$100 & \(\uparrow\) IS \\
        \midrule
        HoloGAN   & 67.5  & 3.95 & 3.52 \\ 
        GRAF      &  41.7 & 2.43 & 3.60 \\ 
        $\pi-$GAN & 29.2 & 1.36 & \bf 4.27 \\
        \midrule
        \our{} (our) & \bf 20.5 & \bf 0.78 & 4.20 \\
        \bottomrule \\
        \end{tabular}
    \caption{Quantitative evaluation of \our{} (our) in comparison with HoloGAN~\cite{nguyen2019hologan},  
GRAF~\cite{schwarz2020graf}, and $\pi$-GAN~\cite{chan2021pi}, in terms of the FID (lower is better), KID (lower is better), and IS (greater is better) metrics. The models were trained on the CARLA dataset~\cite{dosovitskiy2017carla}.
The obtained results demonstrate that our proposed solution is superior (or at least comparable) to the baseline methods.}
    \label{tab:carla_fid_kis_is}    
\end{table}

\begin{table}
    \centering
        \begin{tabular}{cccc}
        \toprule
         CelebA & \(\downarrow\) FID & \(\downarrow\) KID$\times$100 & \(\uparrow\) IS \\
        \midrule
        HoloGAN   & 39.7     & 2.91 & 1.89 \\ 
        GRAF      & 41.1     & 2.29 & 2.34 \\ 
        $\pi$-GAN & \bf 14.7 & \bf 0.39 & 2.62 \\
        \midrule
        \our{} (our) & 15.04 & 0.66 & \bf 2.63 \\
        
        \bottomrule \\
        \end{tabular}
    \caption{Quantitative evaluation of \our{} (our) in comparison with HoloGAN~\cite{nguyen2019hologan},  
GRAF~\cite{schwarz2020graf}, and $\pi$-GAN~\cite{chan2021pi}, in terms of the FID (lower is better), KID (lower is better), and IS (greater is better) metrics. The models were trained on the CelebA dataset~\cite{liu2015deep}.
The obtained results show that our proposed solution achieves similar performance to $\pi$-GAN (the best of the competitors).}
    \label{tab:celeba_fid_kis_is}
\end{table}




\paragraph{3D-aware image synthesis}

In the third experiment, the same models are further compared by modifying the setup to focus on face generation. For this objective, the CelebA dataset~\cite{liu2015deep} is employed, comprising 200,000 high-resolution images of 10,000 different celebrities. The images are cropped from the top of the hair to the bottom of the chin and resized to 128$\times$128 resolution, as in $\pi$-GAN. The quantitative results are presented in Table~\ref{tab:celeba_fid_kis_is}. It is evident that our model and $\pi$-GAN achieve similar performance, which can also be observed in Figure~\ref{fig:celeba_examples}.


\paragraph{Medical applications}

In order to evaluate our model for medical applications and ensure its comparability with existing methods, we utilize a dataset employed by the authors of~\cite{coronafigueroa2022mednerf}, which contains digitally reconstructed radiographs (DRRs). The dataset comprises 20 examples of the chest and 5 examples of the knee, with each example consisting of 72 128$\times$128 images captured at 5-degree intervals, encompassing a full 360-degree vertical rotation for each patient. To account for differences between the synthetic and medical datasets, the sampling angle is modified to encompass a single axis, as in MedNeRF~\cite{coronafigueroa2022mednerf}, and the model configuration is altered to exclude the assumption of a white background for the training data. The model was trained on a single example for each experiment. As illustrated in Figure~\ref{fig:learning}, our qualitative comparison demonstrates that \our{} exhibits a notable enhancement in the quality of reconstruction of CT projections compared to MedNeRF~\cite{coronafigueroa2022mednerf}.
This observation is further substantiated by the quantitative results presented in Table~\ref{tab:fid_kid_scores}.

\section{Conclusions}
    
In this paper, we present \our{}, a novel generative adversarial network (GAN)-based approach to generating 3D-aware representations from 2D images. Our model employs a hypernetwork paradigm and a simplified NeRF representation of a 3D scene. In contrast to the conventional NeRF architecture, \our{} does not utilize viewing directions during training. This enables its successful deployment in diverse datasets where camera position estimation may be challenging or impossible, particularly in the context of medical data. The outcomes of the conducted experiments illustrate that our solution outperforms (or is at least on a comparable level to) existing state-of-the-art methods.

\nocite{langley00}


\newpage
\appendix
\onecolumn
\section{Additional qualitative results of \our{}.}

\begin{figure}[htbp]
    \centering
    \includegraphics[width=0.9\textwidth]{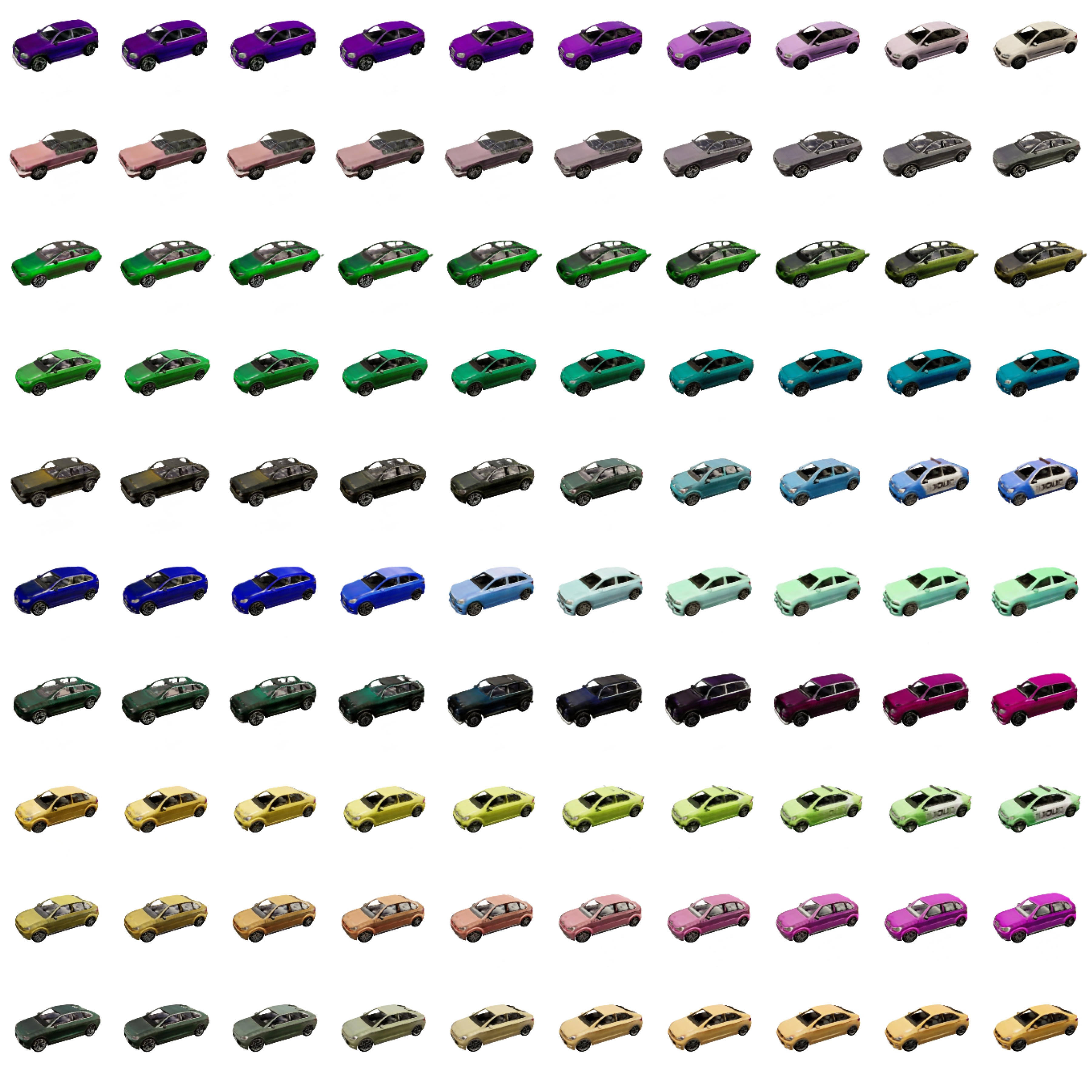}
    \caption{Linear interpolation between latent codes with model trained on CARLA.}
    \label{fig:carla_extra_interpolation_1}
\end{figure}

\begin{figure*}[!h]
    \centering
    \includegraphics[width=0.9\textwidth]{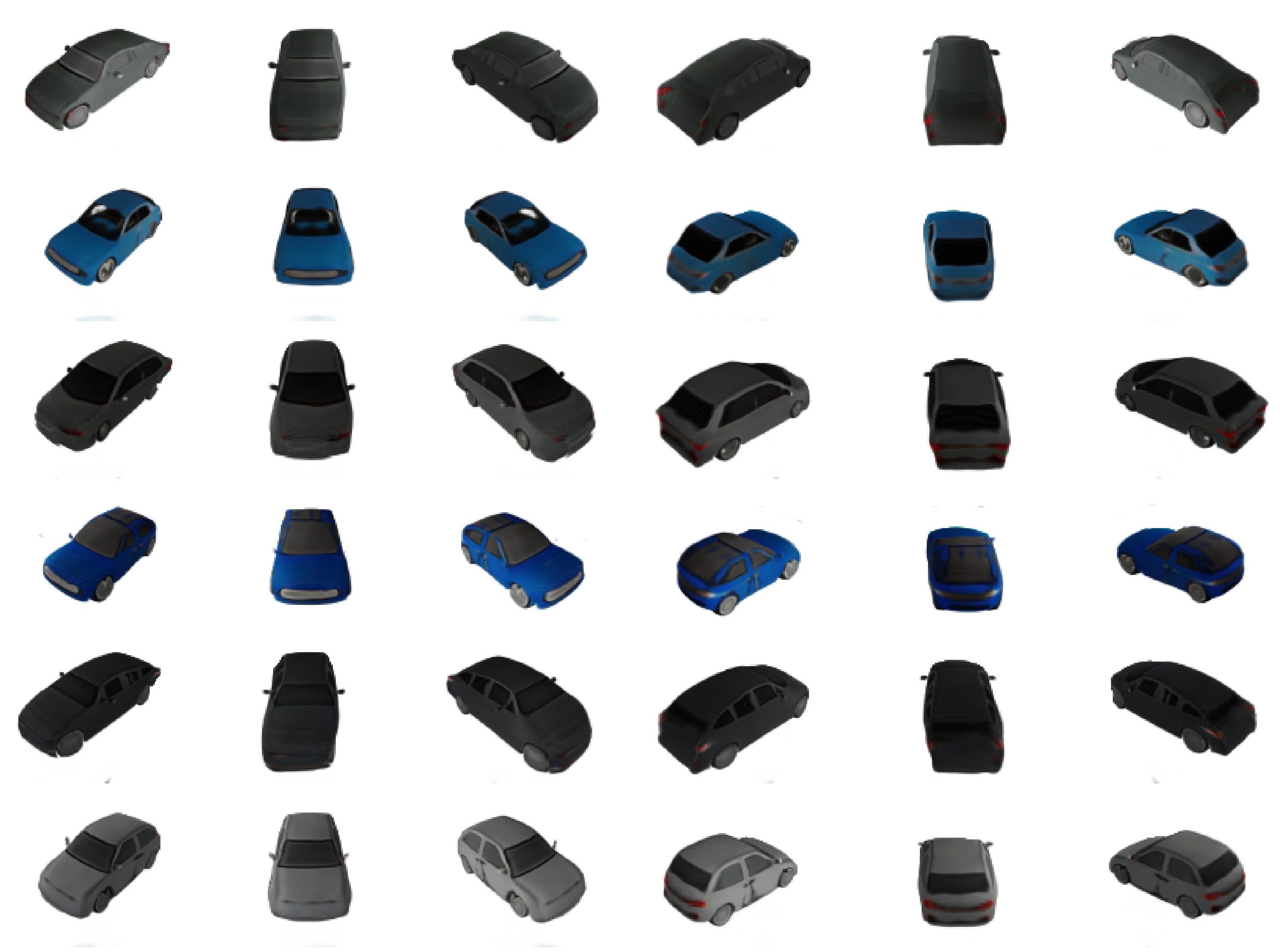}
    \caption{Elements generated by model trained on cars from ShapeNet.}
    \label{fig:shapenet_examples_1}
\end{figure*}

\begin{figure*}[!h]
    \centering
    \includegraphics[width=0.9\textwidth]{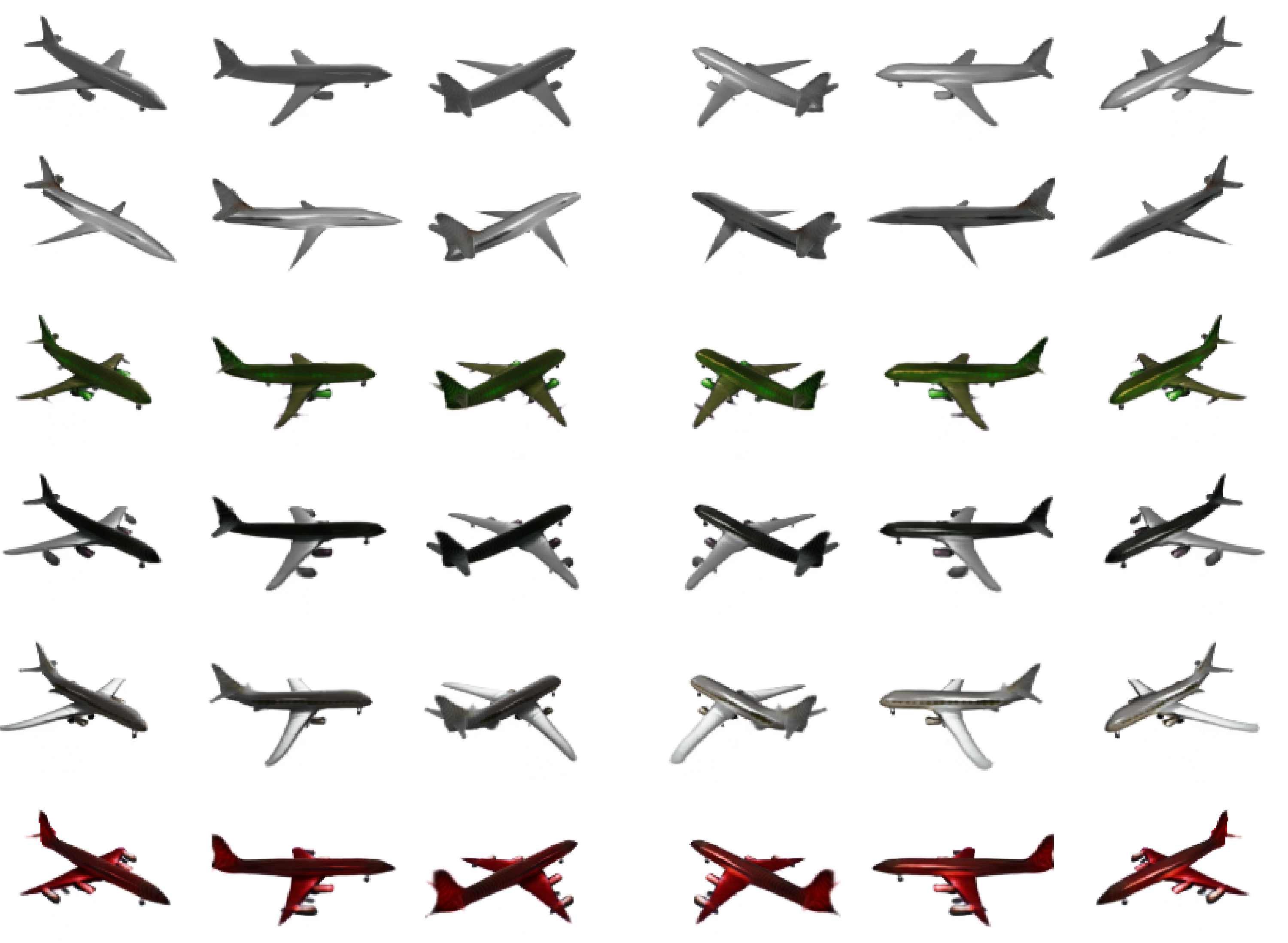}
    \caption{Elements generated by model trained on planes from ShapeNet.}
    \label{fig:shapenet_examples_2}
\end{figure*}


\end{document}